\RequirePackage{fix-cm}
\documentclass[smallextended]{svjour3}       
\smartqed  
\usepackage{graphicx}
\usepackage{changepage}

\usepackage[utf8]{inputenc}

\usepackage{textcomp,marvosym}

\usepackage{fixltx2e}

\usepackage{amsmath,amssymb}

\usepackage{cite}

\usepackage{nameref,hyperref}

\usepackage[right]{lineno}

\usepackage{microtype}
\DisableLigatures[f]{encoding = *, family = * }

\usepackage{rotating}
\usepackage{tablefootnote}

\usepackage{bm,cite,array,fixltx2e,url}

\newcommand{\BigO}[1]{\ensuremath{\operatorname{O}\bigl(#1\bigr)}}
\bmdefine\bmu{\mu}
\bmdefine\bsigma{\sigma}
\bmdefine\bSigma{\Sigma}
\bmdefine\bLambda{\Lambda}
\newcommand{\tab}{\hspace*{0.9em}}
\graphicspath{{img/}}
\DeclareGraphicsExtensions{.pdf,.jpeg,.png}

\usepackage{algorithm}
\usepackage{algorithmic}

\raggedright
\usepackage[aboveskip=1pt,labelfont=bf,labelsep=period,justification=raggedright,singlelinecheck=off]{caption}

\begin{document}

\title{Scalable and Incremental Learning of Gaussian Mixture Models
}


\author{Rafael Coimbra Pinto         \and
        Paulo Martins Engel 
}


\institute{Instituto de Informática, Universidade Federal do Rio Grande do Sul \at
              Av. Bento Gonçalves, 9500 - Agronomia - Porto Alegre, RS - Zip 91501-970 - Brazil \\
              Tel.: +123-45-678910\\
              Fax: +123-45-678910\\
              \email{\{rcpinto,engel\}@inf.ufrgs.br}           
}


\maketitle

\begin{abstract}
This work presents a fast and scalable algorithm for incremental learning of Gaussian mixture models. By performing rank-one updates on its precision matrices and determinants, its asymptotic time complexity is of \BigO{NKD^2} for $N$ data points, $K$ Gaussian components and $D$ dimensions.
The resulting algorithm can be applied to high dimensional tasks, and this is confirmed by applying it to the classification datasets MNIST and CIFAR-10. 
Additionally, in order to show the algorithm's applicability to function approximation and control tasks, it is applied to three reinforcement learning tasks and its data-efficiency is evaluated.

\keywords{Gaussian Mixture Models \and Incremental Learning}
\end{abstract}


\section{Introduction}
	The Incremental Gaussian Mixture Network (IGMN) \cite{heinen2012using,heinen2011igmn} is a supervised algorithm which approximates the EM algorithm for Gaussian mixture models \cite{dempster1977maximum}, as shown in \cite{engel2011incremental}. It creates and continually adjusts a probabilistic model of the joint input-output space consistent to all sequentially presented data, after each data point presentation, and without the need to store any past data points. Its learning process is aggressive,  meaning that only a single scan through the data is necessary to obtain a consistent model.
	
	IGMN adopts a Gaussian mixture model of distribution components that can be 
	expanded to accommodate new information from an input data point, or reduced if spurious components are identified along the 
	learning process. Each data point assimilated by the model contributes to the sequential update of the model parameters based 
	on the maximization of the likelihood of the data. The parameters are updated through the accumulation of relevant 
	information extracted from each data point. New points are added directly to existing Gaussian components or new components are created when necessary, avoiding merge and split operations, much like what is seen in the Adaptive Resonance Theory (ART) algorithms \cite{grossberg1987competitive}. It has been previously shown in \cite{heinen2011connectionist} that the algorithm is robust even when data is presented in random order, having similar performance and producing similar number of clusters in any order. Also, \cite{engel2011incremental} has shown that the resulting models are very similar to the ones produced by the batch EM algorithm.
	
	The IGMN is capable of supervised learning, simply by assigning any of its input 
	vector elements as outputs. In other words, any element can be used to predict any other element, like auto-associative neural networks \cite{rumelhart1986parallel} or missing data imputation \cite{ghahramani1994supervised}. This feature is useful for simultaneous learning of forward and inverse kinematics \cite{damas2012online}, as well as for simultaneous learning of a value function and a policy in reinforcement learning \cite{heinen2011igmnRL}. 
	
	Previous successful applications of the IGMN algorithm include time-series prediction \cite{pinto2011echo,pinto2011recursive,flores2012autocorrelation}, reinforcement learning \cite{heinen2011igmn,heinen2011dealing}, mobile robot control and mapping \cite{pinto2012one,heinen2012using,de2012learning} and outlier detection in data streams \cite{santos2012location}.
	
	However, the IGMN suffers from cubic time complexity due to matrix inversion operations and determinant computations. Its time complexity is of \BigO{NKD^3}, where $N$ is the number of data points, $K$ is the number of Gaussian components and $D$ is the problem dimension. It makes the algorithm prohibitive for high-dimensional tasks (like visual tasks) and thus of limited use. One solution would be to use diagonal covariance matrices, but this decreases the quality of the results, as already reported in previous works \cite{heinen2011connectionist,pinto2011echo}.
	In \cite{pinto2015FIGMN}, we propose the use of rank-one updates for both inverse matrices and determinants applied to full covariance matrices, thus reducing the time complexity to \BigO{NKD^2} for learning while keeping the quality of a full covariance matrix solution.
	
	For the specific case of the IGMN algorithm, to the best of our knowledge, this has not been tried before, although we can find similar efforts for related algorithms. In \cite{Salmen20101903}, rank-one updates were applied to an iterated linear discriminant analysis algorithm in order to decrease the complexity of the algorithm. Rank-one updates were also used in \cite{lefakis2014jointly}, where Gaussian models are employed for feature selection. Finally, in	\cite{olsen2001extended}, the same kind of optimization was applied to Maximum Likelihood Linear Transforms (MLLT).
	
	In this work, we present improved formulas for the covariance matrix updates, removing the need for two rank-one updates, which increases efficiency and stability. It also presents new promising results in reinforcement learning tasks, showing that this algorithm is not only scalable from the computational point-of-view, but also in terms of data-efficiency, promoting fast learning from few data points / experiences.
	
	The next Section describes the algorithm in more detail with the latest improvements to date. Section \ref{sec:figmn} describes our improvements to the algorithm. Section \ref{sec:experiments} shows the experiments and results obtained from both versions of the IGMN for comparison, and Section \ref{sec:conclusion} finishes this work with concluding remarks.

\section{Incremental Gaussian Mixture Network}\label{sec:igmn}

In the next subsections we describe the IGMN algorithm, a slightly improved version of the one described in \cite{pinto2012one}. 
	

\subsection{Learning} \label{sec:learning}
				
	The algorithm starts with no components, which are created as necessary (see subsection \ref{sec:create}). Given input 
	$\textbf{x}$ (a single instantaneous data point), the IGMN algorithm processing step is as follows. First, the squared Mahalanobis distance $d^2(\textbf{x},j)$ for each component $j$ is computed:
	
	\begin{equation}\label{equ:igmn-maha}
		d^2_M(\textbf{x},j) =  (\textbf{x}-\bmu_j)^T \bSigma^{-1}_j (\textbf{x}-\bmu_j)
	\end{equation}

	
	where 
	$\bmu_j$ is the $j^{th}$ component mean, $\bSigma_j$ its full covariance 
	matrix 
	. If any $d^2(\textbf{x},j)$ is smaller than than $\chi^2_{D,1-\beta}$ (the $1-\beta$ percentile of a chi-squared distribution with $D$ degrees-of-freedom, where $D$ is the input dimensionality and $\beta$ is a user defined meta-parameter, e.g., $0.1$)
	, an update will occur, and posterior probabilities are calculated for each component as follows:

	\begin{equation}\label{equ:igmn-like}
		p(\textbf{x}|j) = \frac{1}{(2\pi)^{D/2}\sqrt{|\bSigma_j|}} \exp\left(-\frac{1}{2} d^2_M(\textbf{x},j) \right)
	\end{equation}

	\begin{equation}\label{equ:posterior}
		p(j|\textbf{x}) = \frac{ p(\textbf{x}|j) p(j) }{ \displaystyle\sum\limits_{k=1}^K{ 
		 p(\textbf{x}|k) p(k) } }	\tab \forall{j}
	\end{equation}	

	where $K$ is the number of components. Now, parameters of the algorithm must be updated according to the following equations:
	
	\begin{equation}\label{equ:igmn-v}
		v_j(t) = v_j(t-1) + 1
	\end{equation}

	\begin{equation}\label{equ:igmn-sp}
		sp_j(t) = sp_j(t-1) + p(j|\textbf{x})
	\end{equation}
	
	\begin{equation}\label{equ:igmn-e}
		\textbf{e}_j = \textbf{x} - \bmu_j(t-1)
	\end{equation}
	
	\begin{equation}\label{equ:igmn-omega}
		\omega_j = \frac{ p(j|\textbf{x}) } { sp_j }
	\end{equation}

	\begin{equation}\label{equ:igmn-delta}
		\Delta\bmu_j = \omega_j \textbf{e}_j
	\end{equation}

	\begin{equation}\label{equ:igmn-mu}
		\bmu_j(t) = \bmu_j(t-1) + \Delta\bmu_j
	\end{equation}
	
	\begin{equation}\label{equ:igmn-enew}
		\textbf{e}^*_j = \textbf{x} - \bmu_j(t)
	\end{equation}

	\begin{equation}\label{equ:igmn-C}
		\bSigma_j(t) = (1-\omega_j) \bSigma_j(t-1) + \omega_j \textbf{e}_j^* \textbf{e}_j^{*T} - \Delta\bmu_j \Delta\bmu_j^{T}
	\end{equation}

	\begin{equation} \label{equ:igmn-p}
		p(j) = \frac{ sp_j } { \displaystyle\sum\limits_{q=1}^M{sp_q} }
	\end{equation}
	
	where $sp_j$ and $v_j$ are the accumulator and the age of component $j$, respectively, and $p(j)$ is its prior probability. The equations are derived using the Robbins-Monro stochastic approximation \cite{robbins1951stochastic} for maximizing the likelihood of the model. This derivation can be found in \cite{engel2011incremental,engel2009inbc}.

	
\subsection{Creating New Components} \label{sec:create}
	
	If the update condition in the previous subsection is not met, then a new component $j$ is created and initialized as follows:
			
	\[
		\bmu_j = \textbf{x}; \tab
		sp_j = 1; \tab
		v_j = 1; \tab
		p(j) = \frac{ 1 } { \displaystyle\sum\limits_{i=1}^K{sp_i} }; \tab
		\bSigma_j = \bsigma_{ini}^2 \textbf{I}
	\]
	where $K$ already includes the new component and $\bsigma_{ini}$ can be obtained by:	
	\begin{equation}\label{equ:sigma} 
		\bsigma_{ini} = \delta std(\textbf{x})
	\end{equation}
	
	where $\delta$ is a manually chosen scaling factor (e.g., 0.01) and $std$ is the standard deviation of the dataset. Note that the IGMN is an online and incremental algorithm and therefore it may be the case that we do not have the entire dataset to extract descriptive statistics. In this case the standard deviation can be just an estimation (e.g., based on sensor limits from a robotic platform), without impacting the algorithm.

\subsection{Removing Spurious Components} \label{sec:removing}
	Optionally, a component $j$ is removed whenever $v_j > v_{min}$ and $sp_j < sp_{min}$, where $v_{min}$ and $sp_{min}$ are manually chosen (e.g., 5.0 and 3.0,
	respectively). In that case, also, $p(k)$ must be adjusted for all $k \in K$, $k \ne j$, using (\ref{equ:igmn-p}). In other words, each component is given some time $v_{min}$ to show its importance to the model in the form of an accumulation of its posterior probabilities $sp_j$. Those components are entirely removed from the model instead of merged with other components, because we assume they represent outliers. Since the removed components have small accumulated activations, it also implies that their removal has almost no negative impact on the model quality, often producing positive impact on generalization performance due to model simplification (a more throughout analysis of parameter sensibility for the IGMN algorithm can be found in \cite{heinen2011connectionist}).

\subsection{Inference} \label{sec:recalling}


	In the IGMN, any element can be predicted by any other element. In other words, inputs and targets are presented together as inputs during training. Thus, inference is done by reconstructing data from the target elements
	($\textbf{x}_t$, a slice of the entire input vector $\textbf{x}$) by estimating the posterior probabilities using only the given elements ($\textbf{x}_i$, also a slice of the entire input vector $\textbf{x}$), as follows:

	\begin{equation}\label{equ:recall}
		p(j|\textbf{x}_i) = \frac{ p(\textbf{x}_i|j) p(j) }{ \displaystyle\sum\limits_{q=1}^M{ 
		 p(\textbf{x}_i|q) p(q) } }	\tab \forall{j}
	\end{equation}		
	It is similar to (\ref{equ:posterior}), except that it uses a modified input vector $\textbf{x}_i$ with the target 
	elements $\textbf{x}_t$ removed from calculations. After that, $\textbf{x}_t$ can be reconstructed using the conditional mean equation:
	
	\begin{equation}\label{equ:reconstructfull}
		\hat{\textbf{x}_t} = \displaystyle\sum\limits_{j=1}^M{ p(j|\textbf{x}_i) (\bmu_{j,t} + \bSigma_{j,ti} \bSigma_{j,i}^{-1} (\textbf{x}_i - \bmu_{j,i})) }
	\end{equation}	

	where $\bSigma_{j,ti}$ is the sub-matrix of the $j$th component covariance matrix associating the unknown and known 
	parts of the data, $\bSigma_{j,i}$ is the sub-matrix corresponding to the known part only and $\bmu_{j,i}$ is the $j$th's 
	component mean without the element corresponding to the target element. This division can be seen below:
	
\[
\bSigma_{j} = 
\left (
\begin{array}{r|r}
\bSigma_{j,i} & \bSigma_{j,it} \\
\hline 
\bSigma_{j,ti} & \bSigma_{j,t} \\
\end{array}
\right )
\]

It is also possible to estimate the conditional covariance matrix for a given input, which allows us to obtain error margins for the inference procedure. It is computed according to the following equation:

	\begin{equation}\label{equ:igmn-condconv}
        \hat{\bSigma}(t) = \bSigma_{j,t} - \bSigma_{j,ti}\bSigma_{j,i}^{-1}\bSigma_{j,it}
	\end{equation}	

\section{Fast IGMN}\label{sec:figmn}
In this section, the more scalable version of the IGMN algorithm, the Fast Incremental Gaussian Mixture Network (FIGMN) is presented. It is an improvement over the version presented in \cite{pinto2015FIGMN}.
The main issue with the IGMN algorithm regarding computational complexity lies in the fact that Equation \ref{equ:igmn-maha} (the squared Mahalanobis distance) requires a matrix inversion, which has a asymptotic time complexity of \BigO{D^3}, for $D$ dimensions (\BigO{D^{log_2 7 + \BigO{1}}} for the Strassen algorithm or at best \BigO{D^{2.3728639}} with the most recent algorithms to date \cite{gall2014powers}). This renders the entire IGMN algorithm as impractical for high-dimension tasks. Here we show how to work directly with the inverse of covariance matrix (also called the precision or concentration matrix) for the entire procedure, therefore avoiding costly inversions.

Firstly, let us denote $\bSigma^{-1} = \bLambda$, the precision matrix. Our task is to adapt all equations involving $\bSigma$ to instead use $\bLambda$. 

We now proceed to adapt Equation \ref{equ:igmn-C} (covariance matrix update). This equation can be seen as a sequence of two rank-one updates to the $\bSigma$ matrix, as follows:

	\begin{equation}\label{equ:figmn-C1}
		\bar{\bSigma}_j(t) = (1-\omega_j) \bSigma_j(t-1) + \omega_j \textbf{e}_j^* \textbf{e}_j^{*T}
	\end{equation}

	\begin{equation}\label{equ:figmn-C2}
		\bSigma_j(t) = \bar{\bSigma}_j(t) - \Delta\bmu_j \Delta\bmu_j^{T}
	\end{equation}

This allows us to apply the Sherman-Morrison formula \cite{sherman1950}:

\begin{equation}\label{equ:sherman}
(\textbf{A} + \textbf{uv}^T)^{-1} = \textbf{A}^{-1} - \frac{\textbf{A}^{-1} \textbf{uv}^T \textbf{A}^{-1}}{1 + \textbf{v}^T \textbf{A}^{-1} \textbf{u}}
\end{equation}

This formula shows how to update the inverse of a matrix plus a rank-one update. For the second update, which subtracts, the formula becomes:

\begin{equation}\label{equ:sherman2}
(\textbf{A} - \textbf{uv}^T)^{-1} = \textbf{A}^{-1} + \frac{\textbf{A}^{-1} \textbf{uv}^T \textbf{A}^{-1}}{1 - \textbf{v}^T \textbf{A}^{-1} \textbf{u}}
\end{equation}

In the context of IGMN, we have $\textbf{A} = (1-\omega) \bSigma_j(t-1) = (1-\omega) \bLambda_j^{-1}(t-1)$ and $\textbf{u} = \textbf{v} = \sqrt{\omega}\textbf{e}^*$ for the first update, while for the second one we have $\textbf{A} = \bar{\bSigma}_j(t)$ and $\textbf{u} = \textbf{v} = \Delta\bmu_j$. Rewriting \ref{equ:sherman} and \ref{equ:sherman2} we get (for the sake of compactness, assume all subscripts for $\bLambda$ and $\Delta\bmu$ to be $j$):

\begin{equation}\label{equ:figmn-sherman1}
\bar{\bLambda}(t) = \frac{\bLambda(t-1)}{1-\omega} - \frac{\frac{\omega}{(1-\omega)^2}\bLambda(t-1) \textbf{e}^*\textbf{e}^{*T} \bLambda(t-1)}{1 + \frac{\omega}{1-\omega}\textbf{e}^{*T} \bLambda(t-1) \textbf{e}^*}
\end{equation}

\begin{equation}\label{equ:figmn-sherman2}
\bLambda(t) = \bar{\bLambda}(t) + \frac{\bar{\bLambda}(t) \Delta\bmu\Delta\bmu^T \bar{\bLambda}(t)}{1 - \Delta\bmu^T \bar{\bLambda}(t) \Delta\bmu}
\end{equation}

These two equations allow us to update the precision matrix directly, eliminating the need for the covariance matrix $\bSigma$. They have $\BigO{N^2}$ complexity due to matrix-vector products.

It is also possible to combine the two rank-one updates into one, and this step was not present in previous works. The first step is to combine \ref{equ:figmn-C1} and \ref{equ:figmn-C2} into a single rank-one update, by using equations \ref{equ:igmn-e} to \ref{equ:igmn-enew}, resulting in the following:

	\begin{equation}\label{equ:igmn-C3}
		\bSigma_j(t) = (1-\omega_j) \bSigma_j(t-1) + \textbf{e} \textbf{e}^{T} \omega (1 + \omega (\omega - 3))
	\end{equation}
	
Then, by applying the Sherman-Morrison formula to this new update, we arrive at the following precision matrix update formula for the FIGMN:
	
\begin{equation}\label{equ:figmn-sherman3}
\bLambda(t) = \frac{\bLambda(t-1)}{1-\omega} + \bLambda(t-1) \textbf{e}\textbf{e}^T \bLambda(t-1)
\frac{\omega (1 - 3\omega + \omega^2)}{(\omega-1)^2 (\omega^2 - 2\omega - 1) }
\end{equation}



Although less intuitive than \ref{equ:figmn-C1} and \ref{equ:figmn-C2}, the above formula is smaller and more efficient, requiring much less vector / matrix operations, making FIGMN yet faster and even more stable (\ref{equ:figmn-C2} depends on the result of \ref{equ:figmn-C1}, which may be a singular matrix).

Following on the adaptation of the IGMN equations, Equation \ref{equ:igmn-maha} (the squared Mahalanobis distance) allows for a direct substituion, yielding the following new equation:

	\begin{equation}\label{equ:figmn-maha}
		d^2_M(\textbf{x},j) =  (\textbf{x}-\bmu_j)^T \bLambda_j (\textbf{x} -\bmu_j)
	\end{equation}
	
	which now has a $\BigO{N^2}$ complexity, since there is no matrix inversion as the original equation. Note that the Sherman-Morrison identity is exact, thus the Mahalanobis computation yields exactly the same result, as will be shown in the experiments. After removing the cubic complexity from this step, the determinant computation will be dealt with next.

Since the determinant of the inverse of a matrix is simply the inverse of the determinant, it is sufficient to invert the result. But computing the determinant itself is also a \BigO{D^3} operation, so we will instead perform rank-one updates using the Matrix Determinant Lemma \cite{harville2008matrix}, which states the following:

	\begin{equation}\label{equ:figmn-det-lemma}
	    |\textbf{A} + \textbf{u}\textbf{v}^T| = |\textbf{A}| (1 + \textbf{v}^T \textbf{A}^{-1} \textbf{u})
	\end{equation}
	\begin{equation}\label{equ:figmn-det-lemma2}
	    |\textbf{A} - \textbf{u}\textbf{v}^T| = |\textbf{A}| (1 - \textbf{v}^T \textbf{A}^{-1} \textbf{u})
	\end{equation}

Since the IGMN covariance matrix update involves a rank-two update, adding a term and then subtracting one, both rules must be applied in sequence, similar to what has been done with the $\bLambda$ equations. Equations \ref{equ:figmn-C1} and \ref{equ:figmn-C2} may be reused here, together with the same substitutions previously showed, leaving us with the following new equations for updating the determinant (again, $j$ subscripts were dropped):

	\begin{equation}\label{equ:figmn-det-lemma-apply1}
	    |\bar{\bSigma}(t)| = (1-\omega)^D |\bSigma(t-1)| \left(1 + \frac{\omega}{1 - \omega}\textbf{e}^{*T} \bLambda(t-1) \textbf{e}^*\right)
	\end{equation}

	\begin{equation}\label{equ:figmn-det-lemma-apply2}
	    |\bSigma(t)| = |\bar{\bSigma}(t)| (1 - \Delta\bmu^{T} \bar{\bLambda}(t) \Delta\bmu)
	\end{equation}

Just as with the covariance matrix, a rank-one update for the determinant update is also derived (again, using the definitions from \ref{equ:igmn-e} to \ref{equ:igmn-enew}):

\begin{equation}\label{equ:figmn-det-lemma-apply3}
	    |\bSigma(t)| = (1-\omega)^D |\bSigma(t-1)| 
	    \left(1 + \frac{\omega (1 + \omega (\omega - 3)) }{1-\omega} \textbf{e}^T \bLambda(t-1) \textbf{e} \right)
\end{equation}

This was the last source of cubic complexity, which is now quadratic.

Finishing the adaptation in the learning part of the algorithm, we just need to define the initialization for $\bLambda$ for each component. What previously was $\bSigma_j = \bsigma^2_{ini}\textbf{I}$ now becomes $\bLambda_j = \bsigma^{-2}_{ini}\textbf{I}$, the inverse of the variances of the dataset. Since this matrix is diagonal, there are no costly inversions involved. And for initializing the determinant $|\bSigma|$, just set it to $\prod{\bsigma_{ini}^2}$, which again takes advantage of the initial diagonal matrix to avoid costly operations. Note that we keep the precision matrix $\bLambda$, but the determinant of the covariance matrix $\bSigma$ instead. See algorithms \ref{alg:figmn-learn} to \ref{alg:figmn-create} for a summary of the new learning algorithm (excluding pruning, for brevity).

\begin{algorithm}[ht]
\begin{algorithmic}
\caption{Fast IGMN Learning}\label{alg:figmn-learn}
\REQUIRE{$\delta$,$\beta$,$\textbf{X}$}
\STATE $K > 0$, $\bsigma^{-1}_{ini} = (\delta std(\textbf{X}))^{-1}, M = \emptyset$
\FORALL{input data vector $\textbf{x} \in \textbf{X}$}
    \IF {$K = 0$ \OR $\exists j$, $d^2_M(\textbf{x},j) < \chi^2_{D,1-\beta}$}
        \STATE $update(\textbf{x})$
    \ELSE
        \STATE $M \gets M \cup create(\textbf{x})$
    \ENDIF
\ENDFOR
\end{algorithmic}
\end{algorithm}

\begin{algorithm}[ht]
\begin{algorithmic}
\caption{update}\label{alg:figmn-update}
\REQUIRE{$\textbf{x}$}
\FORALL{Gaussian componentS $j \in M$}

\STATE $d^2_M(\textbf{x},j) =  (\textbf{x}-\bmu_j)^T \bLambda_j (\textbf{x} -\bmu_j)$

\STATE $p(\textbf{x}|j) = \frac{1}{(2\pi)^{D/2}\sqrt{|\bSigma_j|}} \exp\left(-\frac{1}{2} d^2_M(\textbf{x},j) \right)$

\STATE $p(j|\textbf{x}) = \frac{ p(\textbf{x}|j) p(j) }{ \displaystyle\sum\limits_{k=1}^K{ 
		 p(\textbf{x}|k) p(k) } }	\tab \forall{j}$	
\STATE $v_j(t) = v_j(t-1) + 1$

\STATE $sp_j(t) = sp_j(t-1) + p(j|\textbf{x})$
	
\STATE $\textbf{e}_j = \textbf{x} - \bmu_j(t-1)$
	
\STATE $\omega_j = \frac{ p(j|\textbf{x}) } { sp_j }$

\STATE $\bmu_j(t) = \bmu_j(t-1) + \omega_j \textbf{e}_j$
	
\STATE $\bLambda(t) = \frac{\bLambda(t-1)}{1-\omega} + \bLambda(t-1) \textbf{e}\textbf{e}^T \bLambda(t-1)
\frac{\omega (1 - 3\omega + \omega^2)}{(\omega-1)^2 (\omega^2 - 2\omega - 1) }$

\STATE $p(j) = \frac{ sp_j } { \displaystyle\sum\limits_{q=1}^M{sp_q} }$

\STATE $|\bSigma(t)| = (1-\omega)^D |\bSigma(t-1)| 
	    \left(1 + \frac{\omega (1 + \omega (\omega - 3)) }{1-\omega} \textbf{e}^T \bLambda(t-1) \textbf{e} \right)$
	
\ENDFOR
\end{algorithmic}
\end{algorithm}

\begin{algorithm}[ht]
\begin{algorithmic}
\caption{create}\label{alg:figmn-create}
\REQUIRE{$\textbf{x}$}
\STATE $K \gets K + 1$
\RETURN new Gaussian component $K$ with $\bmu_K = \textbf{x}$, $\bLambda_K = \bsigma^{-1}_{ini}\textbf{I}$, $|\bSigma_K| = |\bLambda_K|^{-1}$, $sp_j = 1$, $v_j = 1$, $p(j) = \frac{1}{\displaystyle\sum\limits_{k=1}^K{sp_i}}$
\end{algorithmic}
\end{algorithm}

Finally, the inference Equation \ref{equ:reconstructfull} must also be updated in order to allow the IGMN to work in supervised mode. This can be accomplished by the use of a block matrix decomposition (the $i$ subscripts stand for "input", and refers to the input portion of the covariance matrix, i.e., the dimensions corresponding to the known variables; similarly, the $t$ subscripts refer to the "target" portions of the matrix, i.e., the unknowns; the $it$ and $ti$ subscripts refer to the covariances between these variables):

\begin{equation}\label{equ:blockdecomposition}
\begin{split}
\bLambda_j = \begin{bmatrix}
\bSigma_{j,i} & \bSigma_{j,it} \\
\bSigma_{j,ti} & \bSigma_{j,t}
\end{bmatrix}^{-1} = 
\begin{bmatrix}
\bLambda_{j,i} & \bLambda_{j,it} \\
\bLambda_{j,ti} & \bLambda_{j,t}
\end{bmatrix} \\
= \begin{bmatrix}
(\bSigma_{j,i} - \bSigma_{j,it}\bSigma_{j,t}^{-1}\bSigma_{j,ti})^{-1} & -\bSigma_{j,i}^{-1}\bSigma_{j,it}(\bSigma_{j,t} - \bSigma_{j,ti}\bSigma_{j,i}^{-1}\bSigma_{j,it})^{-1} \\                 -\bSigma_{j,t}^{-1}\bSigma_{j,ti}(\bSigma_{j,i} - \bSigma_{j,it}\bSigma_{j,t}^{-1}\bSigma_{j,ti})^{-1} & (\bSigma_{j,t} - \bSigma_{j,ti}\bSigma_{j,i}^{-1}\bSigma_{j,it})^{-1}  
\end{bmatrix}  
\end{split}
\end{equation}

Here, according to Equation \ref{equ:reconstructfull}, we need $\bSigma_{j,ti}$ and $\bSigma_{j,i}^{-1}$. But since the terms that constitute these sub-matrices are relative to the original covariance matrix (which we do not have), they must be extracted from the precision matrix directly. Looking at the decomposition, it is clear that $\bLambda_{j,it}\bLambda_{j,t}^{-1} = -\bSigma_{j,i}^{-1}\bSigma_{j,it} = -\bSigma_{j,ti}\bSigma_{j,i}^{-1}$ (the terms between parenthesis in $\bLambda_{j,ti}$ and $\bLambda_{j,t}$ cancel each other, while $\bSigma_{j,it} = \bSigma_{j,ti}^T$ due to symmetry). So Equation \ref{equ:reconstructfull} can be rewritten as:

	\begin{equation}\label{equ:figmn-reconstructfull}
		\hat{\textbf{x}_t} = \displaystyle\sum\limits_{j=1}^M{ p(j|\textbf{x}_i) (\bmu_{j,t} - \bLambda_{j,it}\bLambda_{j,t}^{-1} (\textbf{x}_i - \bmu_{j,i})) }
	\end{equation}	

where $\bLambda_{j,it}$ and $\bLambda_{j,t}$ can be extracted directly from $\bLambda$. However, we still need to compute the inverse of $\bLambda_{j,t}$. 
So we can say that this particular implementation has $\BigO{NKD^2}$ complexity for learning and $\BigO{NKD^3}$ for inference. The reason for us to not worry about that is that $d = i + o$, where $i$ is the number of inputs and $o$ is the number of outputs. The inverse computation acts only upon the output portion of the matrix. Since, in general, $o \ll i$ (in many cases even $o = 1$), the impact is minimal, and the same applies to the $\bLambda_{j,it}\bLambda_{j,t}^{-1}$ product. In fact, Weka (the data mining platform used in this work \cite{hall2009weka}) allows for only 1 output, leaving us with just scalar operations.

A new conditional variance formula was also derived to use precision matrices, as it was not present in previous works. Looking again at \ref{equ:igmn-condconv}, we see that it is the Schur Complement of $\bSigma_{j,i}$ in $\bSigma$ \cite{zhang2006schur}. By analysing the block decomposition equation, it becomes obvious that, in terms of the precision matrix $\bLambda$, the conditional covariance matrix has the form:
	\begin{equation}\label{equ:figmn-condconv}
        \hat{\bSigma}(t) = \bLambda_{j,t}^{-1}
	\end{equation}
	
Thus, we are now able to compute the conditional covariance matrix during the inference step of the FIGMN algorithm, which can be useful in the reinforcement learning setting (providing error margins for efficient directed exploration). And better yet, $\bLambda_{j,t}^{-1}$ is already computed in the inference procedure of the FIGMN, which leaves us with no additional computations.

\section{Experiments}\label{sec:experiments}

The first experiment was meant to verify that both IGMN implementations produce exactly the same results. They were both applied to 7 standard datasets distributed with the Weka software (table \ref{datasets}). Parameters were set to  $\delta = 0.5$ (chosen by 2-fold cross-validation) and $\beta = 4.9E-324$, the smallest possible double precision number available for the Java Virtual Machine (and also the default value for this implementation of the algorithm), such that Gaussian components are created only when strictly necessary. The same parameters were used for all datasets. Results were obtained from 10-fold cross-validation (resulting in training sets with 90\% of the data and test sets with the remaining 10\%) and statistical significances came from paired t-tests with $p = 0.05$. As can be seen in table \ref{accuracy}, both IGMN and FIGMN algorithms produced exactly the same results, confirming our expectations. The number of clusters created by them was also the same, and the exact quantity for each dataset is shown in table \ref{clusters}. The Weka packages for both variations of the IGMN algorithm, as well as the datasets used in the experiments can be found at \cite{Pinto2015}. In order to experiment with high dimensional datasets and confirm the algorithm's scalability, it was applied to the MNIST\footnote{\url{http://yann.lecun.com/exdb/mnist/}} and CIFAR-10\footnote{\url{http://www.cs.toronto.edu/~kriz/cifar.html}} datasets as well.

\begin{table}[thb]
\caption{\label{datasets}Datasets}
\scriptsize
{\centering \begin{tabular}{lccc}
\\
\hline
Dataset & Instances (N) & Attributes (D) & Classes \\
\hline
breast-cancer & 286 & 9 & 2 \\
pima-diabetes & 768 & 8 & 2 \\
Glass & 214 & 9 & 7 \\
ionosphere & 351 & 34 & 2 \\
iris & 150 & 4 & 3 \\
labor-neg-data & 57 & 16 & 2 \\
soybean & 683 & 35 & 19 \\
MNIST \cite{lecun1998gradient} & 70000 & 784 & 10 \\
CIFAR-10 \cite{krizhevsky2009learning}  & 60000 & 3072 & 10 \\
\hline
\end{tabular} \scriptsize \par}
\end{table}

Besides the confirmation we wanted, we could also compare the IGMN/FIGMN classification accuracy for the referred datasets against other 4 algorithms: Random Forest (RF), Neural Network (NN), Linear SVM and RBF SVM. The neural network is a parallel implementation of a state-of-the-art Dropout Neural Network \cite{hinton2012improving} with 100 hidden neurons, 50\% dropout for the hidden layer and 20\% dropout for the input layer (this specific implementation can be found at https://github.com/amten/NeuralNetwork). The 4 algorithms were kept with their default parameters. The IGMN algorithms produced competitive results, with just one of them (Glass) being statistically significant below the accuracy produced by the Random Forest algorithm. This value was significantly inferior for all other algorithms too. On average, the IGMN algorithms were the second best from the set, losing only to the Random Forest. Note, however, that the Random Forest is a batch algorithm, while the IGMN learns incrementally from each data point. Also, the resulting Random Forest model used 6 times more memory than the IGMN model. We also tested the FIGMN accuracy on the MNIST dataset, but even after parameter tuning, the results where not on par with the state-of-the-art (above 99\% for deep learning methods), reaching a maximum of around 93\% accuracy. Note, however, that FIGMN is a "flat" machine learning algorithm. Future works may explore the possibility of stacking many levels of FIGMN's in order to obtain better classification results in vision tasks.

\begin{table}[thb]
\caption{\label{accuracy}Accuracy of different algorithms on standard datasets}
\scriptsize
{\centering \begin{tabular}{lr@{\hspace{0cm}}c@{\hspace{0cm}}rr@{\hspace{0cm}}c@{\hspace{0cm}}r@{\hspace{0.1cm}}cr@{\hspace{0cm}}c@{\hspace{0cm}}r@{\hspace{0.1cm}}cr@{\hspace{0cm}}c@{\hspace{0cm}}r@{\hspace{0.1cm}}cr@{\hspace{0cm}}c@{\hspace{0cm}}r@{\hspace{0.1cm}}cr@{\hspace{0cm}}c@{\hspace{0cm}}r@{\hspace{0.1cm}}cr@{\hspace{0cm}}c@{\hspace{0cm}}r@{\hspace{0.1cm}}c}
\\
\hline
Dataset & \multicolumn{3}{c}{RF}& \multicolumn{4}{c}{NN} & \multicolumn{4}{c}{Lin. SVM} & \multicolumn{4}{c}{RBF SVM} & \multicolumn{4}{c}{IGMN} & \multicolumn{4}{c}{FIGMN} \\
\hline
breast-cancer & 69.6 & $\pm$ &  9.1 & 75.2 & $\pm$ &  6.5 &           & 69.3 & $\pm$ &  7.5 &           & 70.6 & $\pm$ & 1.5 &           & 71.4 & $\pm$ & 7.4 &           & 71.4 & $\pm$ & 7.4 &          \\
pima-diabetes & 75.8 & $\pm$ &  3.5 & 74.2 & $\pm$ &  4.9 &           & 77.5 & $\pm$ &  4.4 &           & 65.1 & $\pm$ & 0.4 & $\bullet$ & 73.0 & $\pm$ & 4.5 &           & 73.0 & $\pm$ & 4.5 &          \\
Glass & 79.9 & $\pm$ &  5.0 & 53.8 & $\pm$ &  7.4 & $\bullet$ & 62.7 & $\pm$ &  7.8 & $\bullet$ & 68.8 & $\pm$ & 8.7 & $\bullet$ & 65.4 & $\pm$ & 4.9 & $\bullet$ & 65.4 & $\pm$ & 4.9 & $\bullet$\\
ionosphere & 92.9 & $\pm$ &  3.6 & 92.6 & $\pm$ &  2.4 &           & 88.0 & $\pm$ &  3.5 &           & 93.5 & $\pm$ & 3.0 &           & 92.6 & $\pm$ & 3.8 &           & 92.6 & $\pm$ & 3.8 &          \\
iris & 95.3 & $\pm$ &  4.5 & 95.3 & $\pm$ &  5.5 &           & 96.7 & $\pm$ &  4.7 &           & 96.7 & $\pm$ & 3.5 &           & 97.3 & $\pm$ & 3.4 &           & 97.3 & $\pm$ & 3.4 &          \\
labor-neg-data & 89.7 & $\pm$ & 14.3 & 89.7 & $\pm$ & 14.3 &           & 93.3 & $\pm$ & 11.7 &           & 93.3 & $\pm$ & 8.6 &           & 94.7 & $\pm$ & 8.6 &           & 94.7 & $\pm$ & 8.6 &          \\
soybean & 93.0 & $\pm$ &  3.1 & 93.0 & $\pm$ &  2.4 &           & 94.0 & $\pm$ &  2.2 &           & 88.7 & $\pm$ & 3.0 & $\bullet$ & 91.5 & $\pm$ & 5.4 &           & 91.5 & $\pm$ & 5.4 &          \\
\hline
Average & 85.2 &       &      & 82.0 &       &      &           & 83.1 &       &      &           & 82.4 &       &     &           & 83.7 &       &     &           & 83.7 &       &     &          \\
\hline
\multicolumn{21}{c}{$\bullet$ statistically significant degradation}\\
\end{tabular} \scriptsize \par}
\end{table}

\begin{table}[thb]
\caption{\label{clusters}Number of Gaussian components created}
\scriptsize
{\centering \begin{tabular}{ccc}
\\
\hline
Dataset & \# of Components\\
\hline
breast-cancer & 14.2 $\pm$ 1.9\\
pima-diabetes & 19.4 $\pm$ 1.3\\
Glass & 15.9 $\pm$ 1.1\\
ionosphere & 74.4 $\pm$ 1.4\\
iris &  2.7 $\pm$ 0.7\\
labor-neg-data & 12.0 $\pm$ 1.2\\
soybean & 42.6 $\pm$ 2.2\\
\hline
\end{tabular} \scriptsize \par}
\end{table}

\begin{table}[thb]
\caption{\label{trainingtime}Training and testing running times (in seconds)}
\scriptsize
{\centering \begin{tabular}{ccccc}
\\
\hline
Dataset & IGMN Training & FIGMN Training & IGMN Testing & FIGMN Testing \\
\hline
MNIST &   32,544.69 & 1,629.81 & 3,836.06 & 230.92 \\
CIFAR-10 &   2,758,252* & 15,545.05 & - & 795.98 \\
\hline
\multicolumn{5}{c}{* estimated time projected from 100 data points}\\
\end{tabular} \scriptsize \par}
\end{table}

A second experiment was performed in order to evaluate the speed performance of the proposed algorithm, both the original and improved IGMN algorithms, using the parameters $\delta = 1$ and $\beta = 0$, such that a single component was created and we could focus on speedups due only to dimensionality (this also made the algorithm highly insensitive to the $\delta$ parameter). They were applied to the 2 highest dimensional datasets in table \ref{datasets}, namely, the MNIST and CIFAR-10 datasets. The MNIST dataset was split into a training set with 60000 data points and a testing set containing 10000 data points, the standard procedure in the machine learning community \cite{lecun1998gradient}. Similarlly, the CIFAR-10 dataset was split into 50000 training data points and 10000 testing data points, also a standard procedure for this dataset \cite{krizhevsky2009learning}.

Results can be seen in table \ref{trainingtime}. Training time for the MNIST dataset was ~20 times smaller for the fast version while the testing time was ~16 times smaller. It makes sense that the testing time has shown a bit less improvement, since inference only takes advantage from the incremental determinant computation but not from the incremental inverse computation. For the CIFAR-10 dataset, it was impractical to run the original IGMN algorithm on the entire dataset, requiring us to estimate the total time, linearly projecting it from 100 data points (note that, since the model always uses only 1 Gaussian component during the entire training, the computation time per data point does not increase over time). It resulted in ~32 days of CPU time estimated for the original algorithm against 15545s ($\sim4h$) for the improved algorithm, a speedup above 2 orders of magnitude. Testing time is not available for the original algorithm on this dataset, since the training could not be concluded. Additionally, we compared a pure clustering version of the FIGMN algorithm on the MNIST training set against batch EM (the implementation found in the Weka software). While the FIGMN algorithm took $\sim7.5h$ hours to finish, using 208 Gaussian components, the batch EM algorithm took $\sim1.3h$ to complete a \emph{single iteration} (we set the fixed number of components to 208 too) using 4 CPU cores. Besides generally requiring more than one iteration to achieve best results, the batch algorithm required the entire dataset in RAM. The FIGMN memory requirements were much lower.

Finally, both versions of the IGMN algorithm with $\delta = 1$ and $\beta = 0$ were compared on 11 synthetic datasets generated by Weka. All datasets have 1000 data points drawn from a single Gaussian distribution (90\% training, 10\% testing) and an exponentially growing number of dimensions: 1, 2, 4, 8, 16, 32, 64, 128, 256, 512 and 1024. This experiment was performed in order to compare the scalability of both algorithms. Results for training and testing can be seen in Fig. \ref{figures}:

\begin{figure}[ht]
\centering  \includegraphics[width=0.45\textwidth]{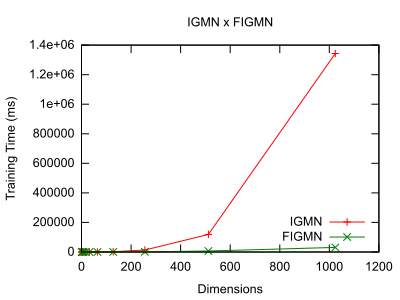}
\includegraphics[width=0.45\textwidth]{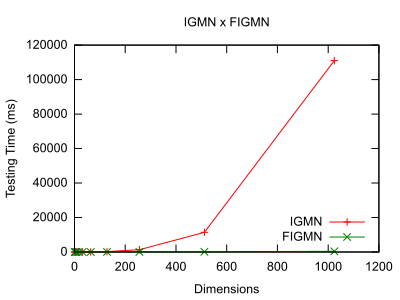}
\caption{Training and testing times for both versions of the IGMN algorithm with growing number of dimensions} \centering \label{figures}

\end{figure}

As predicted, the FIGMN algorithm scales much better in relation to the number of input dimensions of the data.

\subsection{Reinforcement Learning}

Additionally, the FIGMN algorithm was employed for solving three different classical reinforcement learning tasks in the OpenAI Gym \footnote{\url{https://gym.openai.com/algorithms?groups=classic_control}} environment: cart-pole, mountain car and acrobot. Reinforcement learning tasks consist in learning sequences of actions from trial and error on diverse environments.

The mountain car task consists in controlling an underpowered car in order to reach the top of a hill. It must go up the opposite slope to gain momentum first. The agent has three actions at its disposal, accelerating it leftward, rightward,
or no acceleration at all. The agent’s state is made up of two features: current position and speed. The cart-pole task consists in balancing a pole above a small car which can move left or right at each timestep. Four variables are available as observations: current position and speed of the cart and current angle and angular velocity of the pole. Finally, the acrobot task requires a 2-joint robot to reach a certain height with the tip of its "arm". Torque in two directions can be exerted on the 2 joints, resulting in 4 possible actions. Current angle and angular velocity of each joint are provided as observations.

The FIGMN algorithm was compared to other 3 algorithms with high scores on OpenAI Gym: Sarsa($\lambda$), Trust Region Policy Optimization (TRPO; a policy gradient method, suitable to continuous states, actions and time, but which works in batch mode and has low data-efficiency) \cite{schulman2015trust} and Dueling Double DQN (an improvement over the DQN algorithm, using two value function approximators with different update rates and generalizing between actions; it is restricted to discrete actions) \cite{wang2015dueling}. Table \ref{episodes} shows the number of episodes required for each algorithm to reach the required reward threshold for the 3 tasks.

\begin{table}[thb]
\caption{\label{episodes}Number of episodes to solve each task.}
\scriptsize
{\centering \begin{tabular}{ccccc}
\\
\hline
Environment & FIGMN \tablefootnote{\url{https://gym.openai.com/algorithms/alg_xFvqxWu0TSShCouaVW63hg}} & Sarsa($\lambda$) \tablefootnote{\url{https://gym.openai.com/algorithms/alg_hJcbHruxTLOa1zAuPkkAYw}} & TRPO \tablefootnote{\url{https://gym.openai.com/algorithms/alg_yO8abVs8Spm21Icr60SB8g}}   & Duel DDQN \tablefootnote{\url{https://gym.openai.com/algorithms/alg_zy3YHp0RTVOq6VXpocB20g}} \\
\hline
Cart-Pole & 108.80 $\pm$ 22.49  & 557 & 	2103.50 $\pm$ 3542.86 & 	51.00 $\pm$ 7.24 \\
Mountain Car & 	403.83 $\pm$ 79.23 & 1872.50 $\pm$ 6.04 & 4064.00 $\pm$ 246.25 & - \\
Acrobot & 		301.60 $\pm$ 69.12    & 742 & 2930.67 $\pm$ 1627.26 & 31 \\
\hline
\end{tabular} \scriptsize \par}
\end{table}

It is evident that Q-learning with FIGMN function approximation produces better results than Sarsa($\lambda$) with discretized states. Its results are also superior to TRPO's by a large margin. But Duel DDQN appears as the most data-efficient algorithm in this comparison (possibly due to its fixed topology which simplifies the learning procedure).

\section{Conclusion}\label{sec:conclusion}
We have shown how to work directly with precision matrices in the IGMN algorithm, avoiding costly matrix inversions by performing rank-one updates. The determinant computations were also avoided using a similar method, effectively eliminating any source of cubic complexity from the learning algorithm. While previous works used two rank-one updates for covariance matrices and determinants, this work shows how to perform such updates with single rank-one operations. These improvements resulted in substantial speedups for high-dimensional datasets, turning the IGMN into a good option for this kind of tasks. The inference operation still has cubic complexity, but we argue that it has a much smaller impact on the total runtime of the algorithm, since the number of outputs is usually much smaller than the number of inputs. This was confirmed in the experiments.

Reinforcement learning experiments were also useful for showing that the FIGMN algorithm is data-efficient, i.e., it requires few data points in order to learn a usable model. Thus, besides being computationally fast, it also learns fast.

In general, we could see that the fast IGMN is a good option for supervised learning, with low runtimes, good accuracy and high data-efficiency. It should be noted that this is achieved with a single-pass through the data, making it also a valid option for data streams.

\nolinenumbers

\end{document}